\documentclass[sigconf,authorversion,nonacm]{acmart} 

\usepackage{hyperref}
\usepackage{algorithm}
\usepackage{algpseudocode}
\usepackage{bbm}
\usepackage{makecell}
\usepackage{amsfonts}
\usepackage{amsmath}
\usepackage{graphicx}
\usepackage{listings}
\usepackage{booktabs}
\usepackage{tikzducks}
\usepackage{dblfloatfix}

\usepackage{pifont}
\newcommand{\cmark}{\ding{51}}%
\newcommand{\xmark}{\ding{55}}%

\begin{document}

\title{Do Language Models Learn Semantics of Code?\\
A Case Study in Vulnerability Detection}

\author{Benjamin Steenhoek}
\affiliation{%
\institution{Iowa State University}
\city{Ames, Iowa}
\country{USA}
}
\email{benjis@iastate.edu}
\author{Md Mahbubur Rahman}
\affiliation{%
\institution{Iowa State University}
\city{Ames, Iowa}
\country{USA}
}
\email{mdrahman@iastate.edu}
\author{Shaila Sharmin}
\affiliation{%
\institution{Iowa State University}
\city{Ames, Iowa}
\country{USA}
}
\email{ssharmin@iastate.edu}
\author{Wei Le}
\affiliation{%
\institution{Iowa State University}
\city{Ames, Iowa}
\country{USA}
}
\email{weile@iastate.edu}

\newcommand{\discussion}[1]{\textcolor{magenta}{DISCUSS: #1}}
\newcommand{\citeme}{{{\color{red}(needs citation)}}}
\newcommand{\note}[1]{{{\color{red}Note: #1}}}
\newcommand{\xxx}{\textcolor{red}{XXX}}

\newcommand{\wei}[1]{{{\color{blue}Wei: #1}}}
\newcommand{\shaila}[1]{{{\color{purple}Shaila: #1}}}
\newcommand{\md}[1]{{{\color{green}Mahbubur: #1}}}

\newcommand{\ben}[1]{{{\color{orange}#1}}}

\newcommand\getduck{\fpeval{randint(1,100)}}

\keywords{deep learning,vulnerability detection}

\begin{abstract}

Recently, pretrained language models have shown state-of-the-art performance on the vulnerability detection task.
These models are pretrained on a large corpus of source code, then fine-tuned on a smaller supervised vulnerability dataset.
Due to the different training objectives and the performance of the models, it is interesting to consider whether the models have learned the semantics of code relevant to vulnerability detection, namely {\it bug semantics}, and if so, how the alignment to bug semantics relates to model performance.

In this paper, we analyze the models using three distinct methods: interpretability tools, attention analysis, and {\it interaction matrix} analysis.
We compare the models' influential feature sets with the bug semantic features which define the causes of bugs, including buggy paths and {\it Potentially Vulnerable Statements (PVS)}.
We find that (1) better-performing models also aligned better with PVS,
(2) the models failed to align strongly to PVS, and
(3) the models failed to align at all to buggy paths.
Based on our analysis, we developed two annotation methods which highlight the bug semantics inside the model's inputs. We evaluated our approach on four distinct transformer models and four vulnerability datasets and found that our annotations improved the models' performance in the majority of settings -- 11 out of 16, with up to 9.57 points improvement in F1 score compared to conventional fine-tuning.
We further found that with our annotations, the models aligned up to 232\% better to potentially vulnerable statements.
Our findings indicate that it is helpful to provide the model with information of the bug semantics, that the model can attend to it, and motivate future work in learning more complex path-based bug semantics.
Our code and data are available at this URL: \url{https://figshare.com/s/4a16a528d6874aad51a0}.
\end{abstract}

\maketitle

\title{Do Language Models Learn Semantics of Code? A Case Study in Vulnerability Detection} 


\section{Introduction}
Vulnerabilities cause great harm to people and corporations each year~\cite{wikipedia_databreaches_2021,ibm_cost_2021}.
The number of reported vulnerabilities continues to grow~\cite{cvedetails_2021}, which necessitates effective methods of detecting vulnerabilities.
Recently, pre-trained language models have achieved the state-of-the-art vulnerability detection performance of 91 F1 score~\cite{fu_linevul_2022} on a real-world vulnerability dataset~\cite{fan_msr_bigvul_2020}.
However, deep learning models face difficulties in generalizing to new projects and types of bugs and maintaining stable performance~\cite{steenhoek2023empirical}.
These issues make it hard for these models to find vulnerabilities in real-world projects. In this paper, we seek to analyze the models in order to understand the reasons for the models' limitations so that we can find ways to improve them.

At present, pre-trained language models are exclusively trained on textual data and lack explicit integration of program semantics.
Several studies have shown that deep learning-based vulnerability detection models tend to focus on spurious features, e.g., variable names, which are not related to bugs at all~\cite{steenhoek2023empirical,chakraborty_reveal_2022}.
Without leveraging bug semantics, we argue that the model will be unable to reliably and precisely detect bugs.
%
%
For example, consider a snippet code that consists of a buffer allocation \texttt{char buf[10];}, a benign print statement \texttt{printf("\%d\textbackslash{}n",sizeof(buf))}, and an out-of-bounds access \texttt{buf[11];}.
Model A may learn to use spurious features, e.g. the  \texttt{printf} token, or even future unrelated \texttt{printf}, to predict a vulnerability.
Model B, which uses partial bug semantics, may recognize the buffer allocation and access as a potential vulnerability and flag them, but may have a false positive if it cannot further reason whether the index is in-bounds.
Model C, which uses full bug semantics, will follow the buggy paths to track the size of the buffer and check every access to precisely detect overflows. We believe that model C can best detect the vulnerability.
Especially in situations where we cannot obtain a large dataset, such as vulnerability detection, the model's predictive quality hinges on its knowledge of bug semantics.
Therefore, it is crucial to understand whether and how state-of-the-art deep learning models learn bug semantics.

%

Recently, Wan et al. analyzed the self-attention mechanism in transformers and showed that pre-trained models of code can capture and reconstruct basic syntactic structures~\cite{wanWhatTheyCapture2022}.
This builds on work in other domains, which study the relationship of attention with the structures of proteins~\cite{vigBERTologyMeetsBiology2021} and natural language~\cite{jawaharWhatDoesBERT2019}.
In another vein, Paltenghi et al.~\cite{paltenghiExtractingMeaningfulAttention2022} found that the CodeGen transformer model paid attention to similar parts of code as did developers.
These and related research~\cite{rogersPrimerBERTologyWhat2020} show that the models, beyond processing sequences of tokens, learn some high-level properties of the source code.
However, it is not known whether pre-trained models of code can further capture program semantics, especially when fine-tuned for a task which requires the model to learn bug semantics, such as vulnerability detection.
Bug semantics involve more complex constraints and are less related to the text representation compared to the AST~\cite{yamaguchiModelingDiscoveringVulnerabilities2014}, so the model is less likely to learn them by default.

In this paper, we define a set of features related to bug semantics and analyze how deep learning-based language models use these features, in order to answer the following questions:
%
\textit{To what extent do the models use bug root causes to make predictions?
Do models perform better when using more causal lines?}
To this end, we studied four pre-trained language models using interpretation tools, attention analysis~\cite{wanWhatTheyCapture2022}, and interaction matrix analysis~\cite{paltenghiExtractingMeaningfulAttention2022}.
We compared with two types of bug semantics used in the program analysis literature: buggy paths (in this paper, we used paths reported by the Infer static analyzer~\cite{zheng_d2a_2021}; in the future, we will consider other approaches of obtaining buggy paths) and {\it Potentially Vulnerable Statements (PVS)}, based on the framework proposed by Le et al.~\cite{le_marple_2008}.
We found that the models aligned to the PVS in some cases, but most models had less than 50\% alignment with potentially vulnerable statements. However, the models did not substantially align to buggy paths, having median alignment scores even below 7\%.
Based on this analysis, we developed two methods to annotate the PVS inside the model input. Our best annotation method improved the model F1 score in the majority of cases, by up to 9.57 points, and improved model alignment to PVS by 36-232\%.

In this paper, we make the following contributions:
\begin{enumerate}
    \item We adapt and harmonize three distinct approaches to analyze deep learning models using interpretation tools, attention analysis, and interaction matrix analysis.
    \item Using these analysis approaches, we investigate whether deep learning models encode two types of bug semantics: Potentially Vulnerable Statements (PVS) and buggy paths.
    \item We show that better-performing models aligned better to bug semantics, and while the models somewhat aligned with PVS, they failed to align strongly, and failed to align substantially to buggy paths.
    \item Therefore, we develop a novel method of annotating bug semantics and show that with this annotation, models can perform significantly better and indeed aligned more with PVS.
\end{enumerate}

\section{Background}
\label{sec:background}

\subsection{Vulnerability detection}

A vulnerability is a defect in a program which can be exploited for harm to the program's user or system.
These can be caused by erroneous operations such as memory allocations, pointer manipulation, lack of input validation.
We studied vulnerability detection in the setting of classifying function-level source code examples into vulnerable or non-vulnerable.

Critically, vulnerabilities can be analyzed and detected according to conditions under which a vulnerability manifests~\cite{yamaguchiModelingDiscoveringVulnerabilities2014}, which we term as \textit{bug semantics}.
Static vulnerability analyzers utilize bug semantics, such as buggy paths with pre- and post-conditions, to identify likely security vulnerabilities in software source code.
Recent transformer models can outperform prior approaches, but these models rely on textual representations which do not directly represent bug semantics, leading us to ask: how do these models perform so well, do they really learn bug semantics, and can they be improved?


\subsection{Transformers and self-attention}

The transformer model architecture, introduced by Vaswani et al.~\cite{vaswani_attention_2017}, has recently revolutionized the field of text processing, including source code understanding.
The key innovation of the transformer model is that it relies solely on the self-attention mechanism without using recurrent neural networks (RNNs) or convolutions.
This architecture allows greater computational efficiency, allowing it to scale to large parameter sizes and training corpora.

Self-attention relates different elements of a sequence $c = [w_1, \ldots, w_n]$ of size $n$ in order to compute a representation of the sequence.
The transformer model is made up of an embedding layer and $L$ sequential self-attention layers.
The embedding converts the sequence of tokens into vector representations $x^0 = [h^0_1, \ldots, h^0_n]$.
Each attention layer $l$ takes the previous layer's output and generates a vector representation $H^l = [h^l_1, \ldots, h^l_n]$.
Each attention layer has multiple self-attention heads which each compute a separate output representation.
Multiple heads allow the model to focus on different tokens, resulting in improved performance; in fact, sometimes individual heads will tend to align with a semantic function, as shown by Wan et al.~\cite{wanWhatTheyCapture2022}.
Intuitively, a higher score for an index $j$ means that more of the previous layer's representation of token $j$ will be preserved.
To compute the attention in parallel, most implementations use the vectorized formula in Equation \ref{eq:attention}.

\begin{center}
    \label{eq:attention}
    \begin{align}
        Attention(Q, K, V) = softmax(\frac{Q K^T}{\sqrt{d}}) \cdot V
    \end{align}
\end{center}

where $d$ is a hyperparameter specifying the dimension of the model's hidden representation and Q, K, and V are linear mappings of the previous hidden representation.
The output of the softmax is the attention score which scales the value of $V$; a high score for a token $j$ will mostly preserve the previous layer's representation of $j$, and a low score will make the representation close to zero.
The output of the attention head is summed, then concatenated with the other attention heads and projected, and sent to a feed-forward neural network to produce the output of the layer.



To produce the input sequence $c$, text input sequences are commonly split into \textit{input tokens} using a Byte Pair Encoding (BPE)~\cite{sennrichNeuralMachineTranslation2016}. This tokenization technique is used to address the out-of-vocabulary problem by splitting words into common fragments -- ``subwords'' -- allowing the model to generalize to unseen vocabulary.


\subsection{Pre-trained source code models}
Transformer models are especially effective when they are pretrained on large corpora of text and source code, then fine-tuned for specific tasks.
The most common BERT-based models are usually trained with the Masked Language Modeling (MLM) objective, where the model takes as input a sequence of tokens $c$, a fraction of the tokens are hidden, and the model is tasked with filling them in.
Pre-training can integrate other inputs and objectives, such as including ASTs in the input or tasking the model with identifier reconstruction~\cite{wang_codet5_2021,guo_unixcoder_2022}.
Surprisingly, the models can still learn high-level patterns from the code during pre-training in order to fulfill the MLM objective.
%

Recently, Wan et al.~\cite{wanWhatTheyCapture2022} have shown that pre-trained source code models can encode the syntactic structure of the code.
We apply and extend the attention analysis technique they introduced in order to compare instead with bug semantics. Rather than studying the pre-trained models, we studied the models after fine-tuning them for vulnerability detection, in order to understand if they learned the bug semantics.

To fine-tune the model for vulnerability detection, we load the pre-trained weights and replace the MLM prediction layer with a feed-forward classification layer.
Then we fine-tune the model on a smaller binary classification dataset for vulnerability detection, following the procedure introduced by Fu et al.~\cite{fu_linevul_2022}.

\newcommand{\gt}{B}         
\newcommand{\g}{b}          
\newcommand{\mi}{M}         
\newcommand{\gtedge}{\gt_\times}  
\newcommand{\gtpath}{\gt_{walk}}  
\newcommand{\miedge}{M_\times}  
\newcommand{\astree}{T}     
\newcommand{\ds}{D}         
\newcommand{\im}{IM}         

\section{Study Setup}
\label{sec:setup}

\subsection{Models}
We studied the following SOTA pre-trained language models:
%
%
\textbf{CodeBERT}~\cite{feng_codebert_2020} is a model pretrained on pairs of source code and natural language. It is an early and common baseline model~\cite{lu_codexglue_2021} upon which other works have built more advanced approaches.
\textbf{LineVul}~\cite{fu_linevul_2022} is based on the CodeBERT backbone and, furthermore, produces line-level explanations based on self-attention.
\textbf{UniXCoder}~\cite{guo_unixcoder_2022} incorporates AST and code comments into its model input and adds pretraining tasks such as contrastive learning and PL$\rightarrow$NL comment generation.
\textbf{CodeT5}~\cite{wang_codet5_2021} annotates identifiers in its model input and adds identifier-aware and dual NL$\rightarrow$PL and PL$\rightarrow$NL generation pretraining tasks.

\subsection{Datasets}

We fine-tuned the models on the following datasets of bugs in C and C++ source code, which we also use later in our studies.
We chose these datasets because they are widely used and represent diverse dataset collection methods. Table \ref{tab:dataset-information} lists the details of each dataset.
\textbf{D2A}~\cite{zheng_d2a_2021} used the Infer static analyzer and differential analysis to collect buggy and the corresponding fixed functions.
We used the function-level leaderboard dataset\footnote{\url{https://developer.ibm.com/exchanges/data/all/d2a/}}, which is balanced.
\textbf{Devign}~\cite{zhou_devign_2019} used commit filtering and further manual analysis to gather a balanced dataset of buggy and non-buggy functions.
We used the partitions released by the CodeXGLUE dataset \cite{lu_codexglue_2021}.
\textbf{Big-Vul}~\cite{fan_msr_bigvul_2020} crawled the Common Vulnerabilities and Exposures (CVE) database to collect an imbalanced dataset of buggy and non-buggy functions.
We used the partitions released by Fu et al.~\cite{fu_linevul_2022}.
\textbf{ReVeal}~\cite{chakraborty_reveal_2022} crawled public bug repositories for Chrome and Debian to produce an imbalanced dataset of buggy/fixed functions and unrelated functions in the same commit, which are assumed non-vulnerable.

\begin{table}[htb]
    \centering
    \caption{Dataset information}
    \label{tab:dataset-information}
    \begin{tabular}{lrrcc}
        \toprule
        \thead{Dataset} & \thead{\# data} & \thead{\% vulnerable} & \thead{Buggy paths} & \thead{PVS} \\\hline
        D2A~\cite{zheng_d2a_2021}     & 5,239   & 53.35\%   & \cmark & \cmark \\\hline
        Devign~\cite{zhou_devign_2019}  & 27,318  & 45.61\%   & \xmark & \cmark \\\hline
        Big-Vul~\cite{fan_msr_bigvul_2020} & 188,636 & 5.78\%  & \xmark & \cmark \\\hline
        ReVeal~\cite{chakraborty_reveal_2022}  & 22,734  & 9.85\% & \xmark & \cmark \\
        \bottomrule
    \end{tabular}
\end{table}

We report the performances of the reproduced models fine-tuned on each dataset in Table \ref{tab:reproduced-performance}.
For fair comparison, we compared all model architectures using a context length of 512 in all of our experiments.
In order to give the same inputs to all the models, we also normalized the whitespace inside the programs by joining all AST tokens with a single space; this does not change the semantics of the C/C++ programs, aside from exceptional cases like preprocessor macros.

\begin{table}[htb]
    \centering
    \caption{Reproduced model performance}
    \label{tab:reproduced-performance}
\begin{tabular}{lrrrr}
\toprule
& \multicolumn{4}{c}{F1 score} \\\cline{2-5}
\thead{Model}         & \thead{D2A}   & \thead{Devign} & \thead{Big-Vul}    & \thead{ReVeal} \\\hline
CodeBERT      & 66.76 & 56.90  & 40.65  & 42.69 \\\hline
UniXcoder     & 57.19 & 56.81  & 39.55  & 40.53 \\\hline
CodeT5        & 57.33 & 58.79  & 40.20  & 40.56 \\\hline
LineVul       & 68.22 & 54.15  & 39.46  & 42.92 \\
\bottomrule
\end{tabular}
\end{table}





\newcommand{\myemph}[1]{\underline{#1}}

\subsection{Computing bug semantics}

In order to identify the statements which are relevant to the bug, 
we define 
\textit{bug semantic features}, or {\it bug features}, as elements from the source code which cause a bug; once these elements are eliminated, the bug ceases to exist.
We hypothesize that reliable and robust bug detection models should concentrate on bug semantic features to make their predictions.
For example, when detecting a buffer overflow, a model should track buffer size and string length, and detecting memory leak, we should determine if a {\tt free} is performed after a memory allocation.
If the models overlook the root cause of the bug, the prediction cannot be reliable and robust~\cite{yamaguchiModelingDiscoveringVulnerabilities2014}.

In Section~\ref{sec:correlation}, we will analyze the models' important features and attention, comparing them with the bug semantic features. This will help us understand \textit{whether} the models learn to concentrate on bug semantics to make predictions (\S\ref{sec:correlation}) and \textit{how} we can leverage bug semantics to improve the models (\S\ref{sec:improvement}).


We extract bug features using a lightweight static analysis on a program's Abstract Syntax Tree (AST).
Given a dataset of functions $D$, we parse each function into an AST $\astree = \langle V, E \rangle$.
\textit{Terminal} nodes are the nodes in $V$ which have no children.
We then extract two types of bug features as follows:

\paragraph{Buggy paths}

A \textit{buggy path} consists of a sequence of statements in source code that lead to a buggy condition. The D2A dataset directly used the buggy path reported by the Infer static analyzer as the ground-truth to distinguish buggy and non-buggy functions~\cite{zheng_d2a_2021}.
Therefore, in our static analysis tool, we compute a buggy path as an ordered set of terminal AST nodes in $V$ which are inside the lines identified in the buggy path reported by Infer. Note that it is undecidable to automatically extract ground-truth buggy paths that track the root cause a bug. Here, we use the buggy paths reported by Infer as an approximation.
Since the static analysis tools are robust and accurate enough to be used widely in industry, we expect that accurate vulnerability detection models should make decisions based on many of the same features. In fact, the buggy paths were used to generate the labels for the D2A dataset, so when training on D2A, the models have a high incentive to directly focus on the buggy paths.


\paragraph{Potentially Vulnerable Statements (PVS)}
To further study characteristics of bugs, we identified \textit{Potentially Vulnerable Statements (PVS)} -- statements where a vulnerability condition can manifest, following the framework proposed by Le et al. ~\cite{le_marple_2008}.
To select the PVS, we surveyed common buggy function calls and operations in C~\cite{moshtari_grounded_2022,bian_sinkfinder_2020,the_mitre_corporation_cwe_2022} and categorized them based on the types of bugs which they might cause.
Table \ref{tab:potentially-vulnerable-statements} lists the function calls and operations which we considered to indicate a PVS.
In our static analysis tool, we compute PVS as an unordered set of terminal AST nodes in $V$ which are descendants of the selected PVS statement.
This is a heuristic which is insufficient to detect bugs by itself; it can include statements which do not really cause a bug, or exclude statements which do cause bugs. PVS would usually be followed up by a more precise static analyzer, but indicates the very simplest level of bug semantics which we intend the model to align to; we intend for the model to be able to use PVS as a starting point for a more precise analysis.

\begin{table}[htb]
    \centering
    \caption{List of Potentially Vulnerable Statements (PVS)
    }
    \label{tab:potentially-vulnerable-statements}
    \begin{tabular}{c c}
        \toprule
        \thead{Statement} & \thead{Vulnerability Type} \\\midrule
        \makecell{
            Call to {\tt malloc}, {\tt calloc},\\
            {\tt realloc}, {\tt aligned\_alloc},\\
            {\tt kalloc}, {\tt kcalloc}, {\tt krealloc},\\
            {\tt valloc}, {\tt vcalloc}, {\tt vrealloc}
        } & Memory leak \\\hline
        \makecell{
            Call to {\tt free}, {\tt kfree},\\
            \texttt{free\_sized},\\\texttt{free\_aligned\_sized}
        } &
        \makecell{Double free, memory leak, \\ use after free}\\\midrule
        \makecell{
            Call to {\tt gets}, {\tt puts}, {\tt scanf},\\
            {\tt sprintf}, {\tt strcpy}, {\tt strncpy},\\
            {\tt strlen}, {\tt strcat}, {\tt strncat}\\
            Array index ({\tt buf[]})
        } & Buffer overflow \\\hline
        \makecell{
            Pointer dereference (\texttt{*ptr})\\
            Pointer member access\\(\texttt{ptr->field})
        } & {\tt NULL} pointer dereference \\\hline
        \makecell[c]{{\tt +}, {\tt +=}, \tt{ ++}, {\tt -},\\\tt{ -=}, \tt{ --}, {\tt *}, \tt{ *=}} & Integer overflow/underflow \\\midrule
        {\tt /, /=, \%, \%=} & Divide-by-zero \\
        \bottomrule
    \end{tabular}
\end{table}

Table \ref{tab:pvs-statistics} lists the frequencies of PVS in the datasets.
The PVS are more frequent in vulnerable examples than non-vulnerable examples in all datasets except D2A, though the non-vulnerable examples have PVS in them; this shows that PVS is necessary but not sufficient conditions for vulnerability. For example, when a program contains a good bounds-check, the buffer access (PVS) is safe. We speculate that the D2A dataset has many PVS in its non-vulnerable examples because it is collected from differentiating bugs reported by the Infer static analyzer with the corresponding fixed version.
Due to its differential data collection, the D2A dataset can log multiple buggy paths per method, though the examples in the evaluation dataset are not duplicated; on average, each program has 5.29 overlapping buggy paths. We count all buggy paths as separate bugs when calculating the alignment metrics.

\begin{table}[htb]
    \centering
    \caption{Statistics of PVS in the datasets.}
    \label{tab:pvs-statistics}
    \begin{tabular}{lrrrr}
    \toprule
         & \multicolumn{4}{c}{Mean \# PVS per program} \\\cline{2-5}
        \thead{Label} & \thead{D2A} & \thead{Devign} & \thead{Big-Vul} & \thead{ReVeal} \\\midrule
        Vulnerable      & 111.1	& 103.6	& 146.2	& 85.5  \\
        Non-vulnerable  & 111.4	& 85.8	& 50.3	& 27.4  \\
        Vul:Non-vul ratio           & 1.00	& 1.21	& 2.90	& 3.12  \\
    \bottomrule
    \end{tabular}
    
\end{table}




\section{Alignment of Model Features with Bug Semantics}
\label{sec:correlation}


In this section, we analyze the \textit{alignment} between the models' important features/attention and bug semantic features.
Our goal is to understand whether the models learned to make predictions based on bug semantic features, rather than unrelated spurious features.
To this end, we corroborated three distinct approaches.

\begin{enumerate}
\item \textbf{Interpretation analysis:}
We applied state-of-the-art interpretability tools to attribute the importance of each feature, then measured the agreement between features with high attribution scores and bug semantic features.

\item \textbf{Attention analysis:}
We measured the agreement between the model's self-attention scores and bug semantic features, adapting the method established by Wan et al.~\cite{wanWhatTheyCapture2022} and Vig et al.~\cite{vigBERTologyMeetsBiology2021}.

\item \textbf{Interaction matrix analysis:}
We compared the features which the model's attention mechanism was likely to focus on in sequence with the bug semantic features, using the interaction matrix proposed by Paltenghi et al.~\cite{paltenghiExtractingMeaningfulAttention2022}.
\end{enumerate}

For a single program AST $T = \langle V, E \rangle$, we denote the bug semantic features as a set $\gt \subseteq V$ and denote the features identified as important by an analysis as a set $\mi \subseteq V$.
We used this shared interface to corroborate the results from each approach.
Each alignment metric produces a scalar score which quantifies the agreement between $\mi$ and $\gt$. For all analyses, we used the Intersection-over-Union (IoU) index to measure the similarity between $M$ and $B$; this controls for the size of $B$, which varies between different programs and datasets.

We fine-tuned each model on each dataset and chose the best checkpoint based on validation performance to compute the alignment metrics; Table \ref{tab:reproduced-performance} reports the model performance comparison.
When computing the metrics, we considered only the programs where $\gt \neq \emptyset$, i.e. which contained at least one PVS or a buggy path.


\begin{figure*}[t]
\centering
\includegraphics[width=0.3\textwidth]{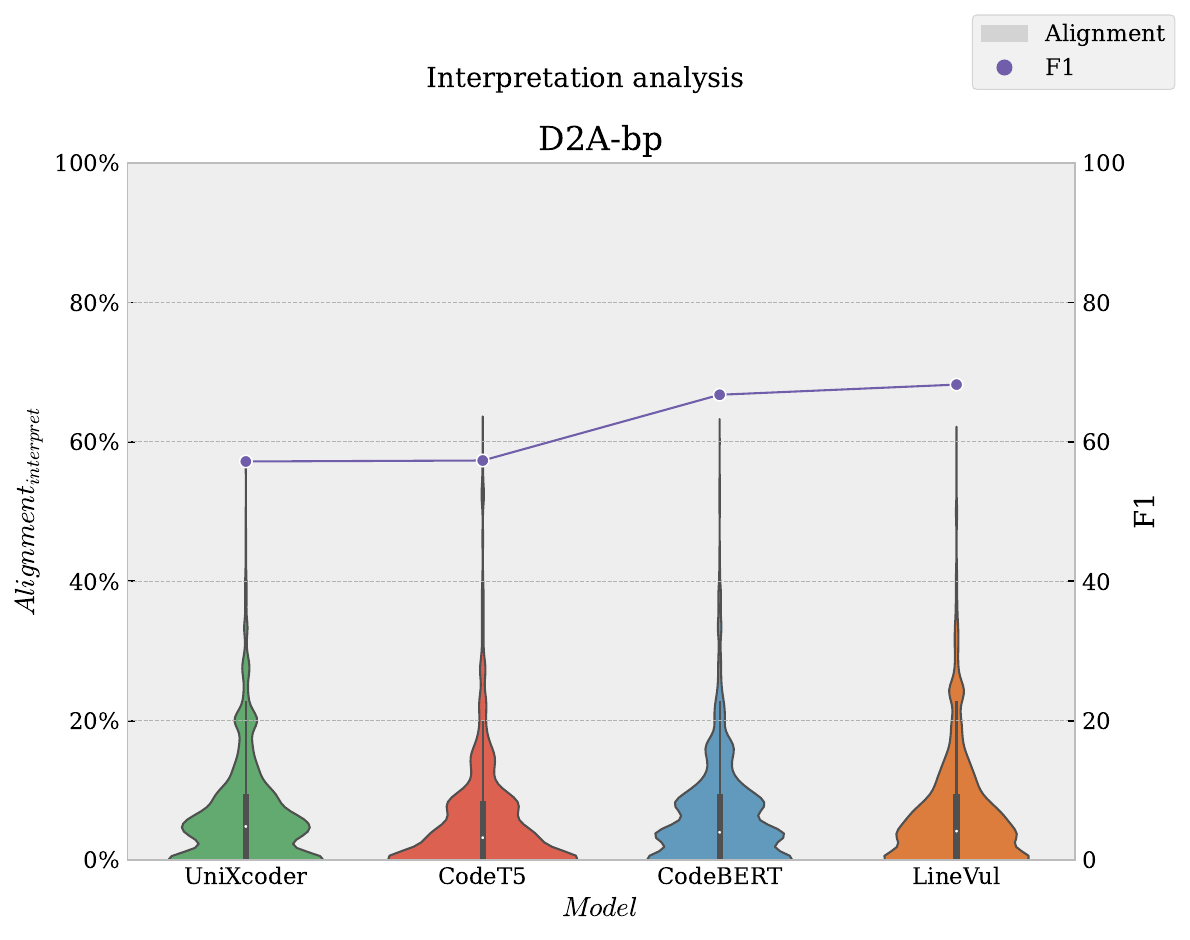}
\includegraphics[width=0.3\textwidth]{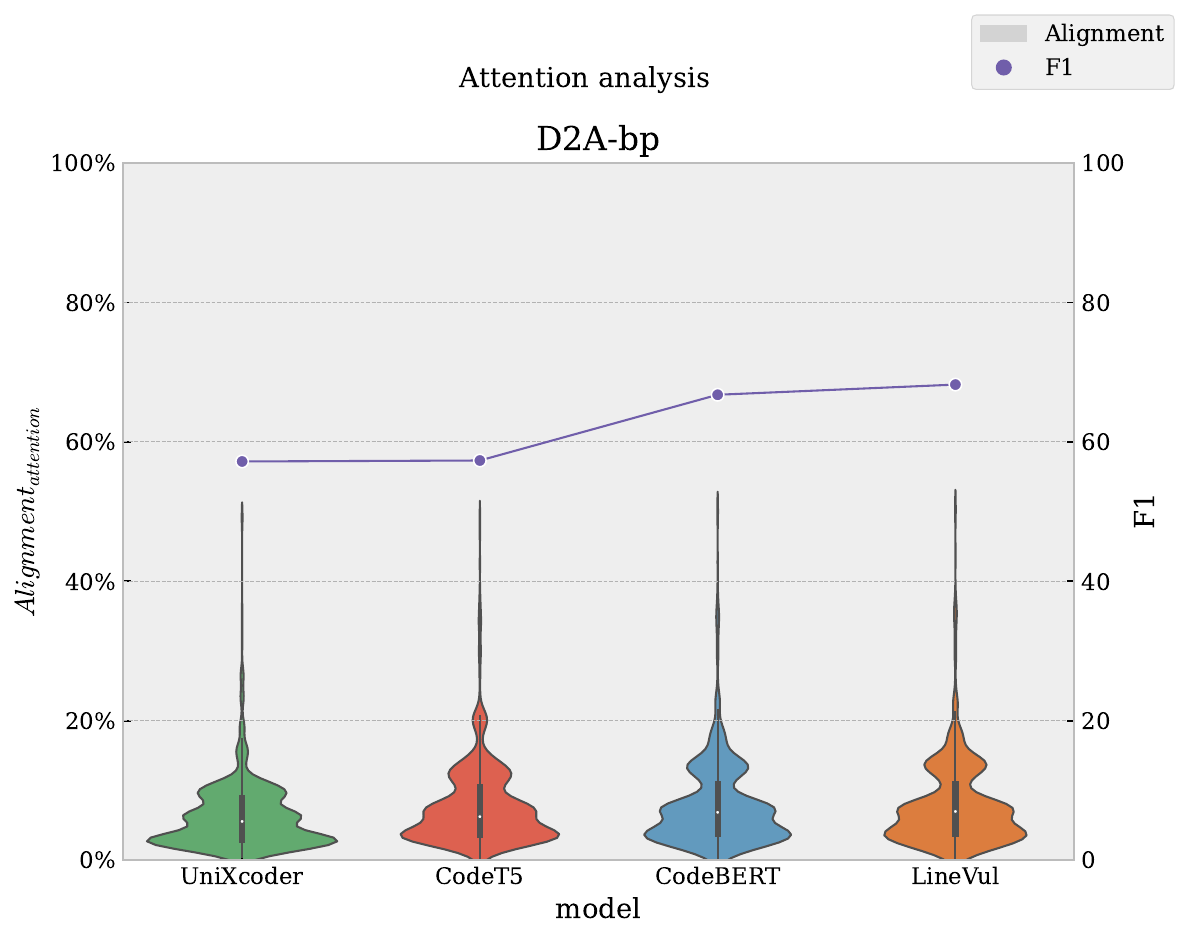}
\includegraphics[width=0.3\textwidth]{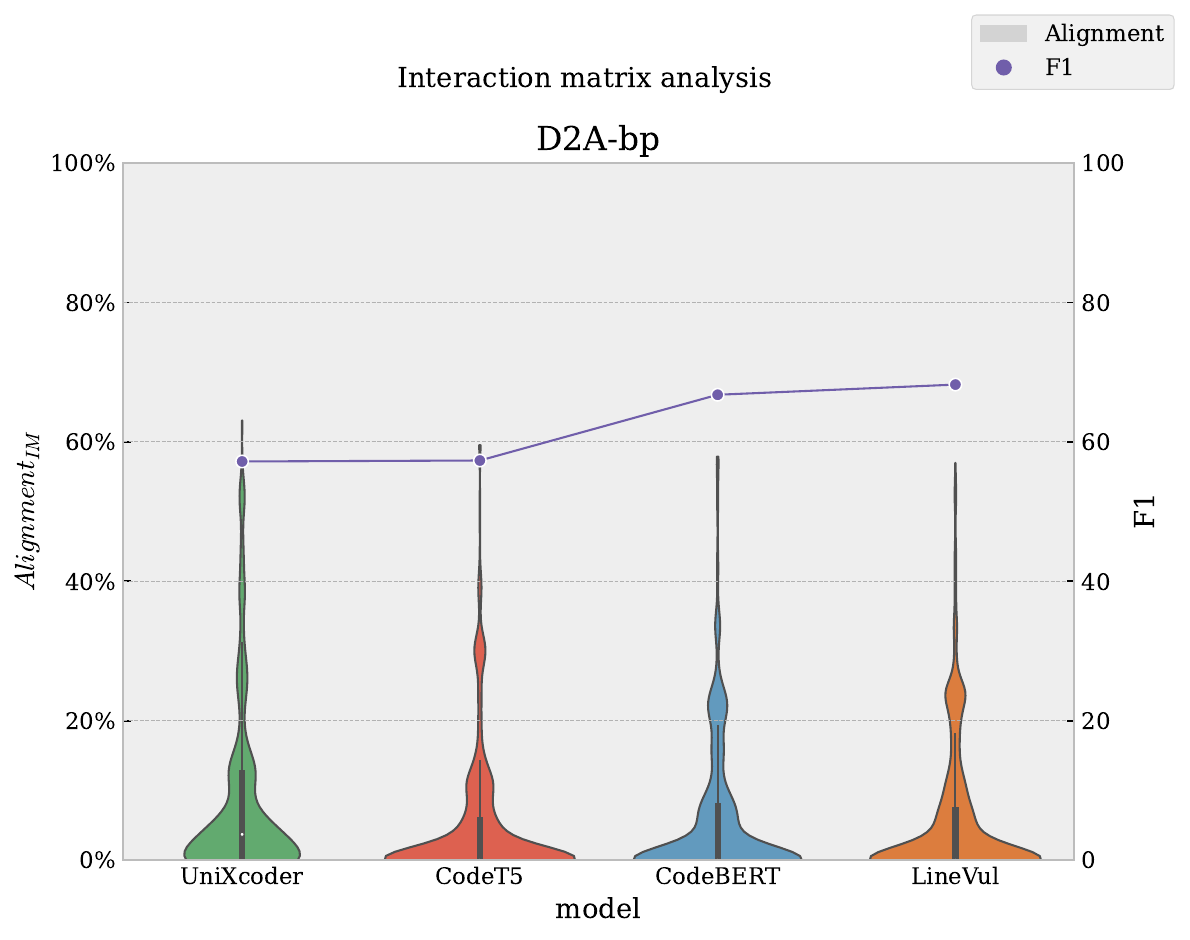}
\caption{Models did not align to D2A buggy paths.
}
\label{fig:d2a-bp}
\end{figure*}

\subsection{Interpretation analysis}

\subsubsection*{Motivation}
Many bugs can only be detected reliably if certain individual statements are considered.
For example, a buffer overflow detector which ignores the statements containing buffer accesses cannot precisely determine whether a buffer overflow may occur.
Although it may make some correct predictions based on heuristics or spurious signals, its predictions will always have additional uncertainty arising solely from oversight of the fault location.
Using interpretation tools, we seek to detect whether the existing models consider \textit{individual} bug features in their predictions.

\subsubsection*{Approach}

Many methods have recently been proposed to generate \textit{interpretations} of model predictions in the form of feature attribution scores. These tools generate a scalar score for each feature (in our case, input tokens) which indicates the feature's importance for the model's prediction.
Because different interpretation tools can yield diverse insights and can be difficult to compare~\cite{doshi-velezRigorousScienceInterpretable2017a}, we applied four state-of-the-art interpretation tools: Saliency, InputXGradient, DeepLift, and SHAP ~\cite{simonyanDeepConvolutionalNetworks2014,shrikumarNotJustBlack2017,lundbergUnifiedApproachInterpreting2017,shrikumarLearningImportantFeatures2019} which were available in the versatile Captum library~\cite{kokhlikyan2020captum}.
We present our IoU formula in Equation \ref{eq:interpretation-analysis}, where $\mi_{topInterpret(k)}$ represents the set of tokens with the top-$k$ highest attribution scores in the program, and $k = |B|$.
Finally, we averaged the results of $Alignment_{Interpret}$ for each tool to produce a single score for each program.

\begin{equation}\label{eq:interpretation-analysis}
Alignment_{Interpret} =
    \frac{|\mi_{topInterpret(k)} \cap \gt|}
    {|\mi_{topInterpret(k)} \cup \gt|}
\end{equation}

Because the interpretability tools operate at the input token-level, we cannot directly compare them with the bug features, which are highlighted at AST token-level. In order to compare $\mi$ and $\gt$ at the same granularity, we averaged the attribution scores within each AST token to produce a single score for each token.

\subsubsection*{Results}

\begin{figure}[b]
\centering
\includegraphics[width=0.49\textwidth]{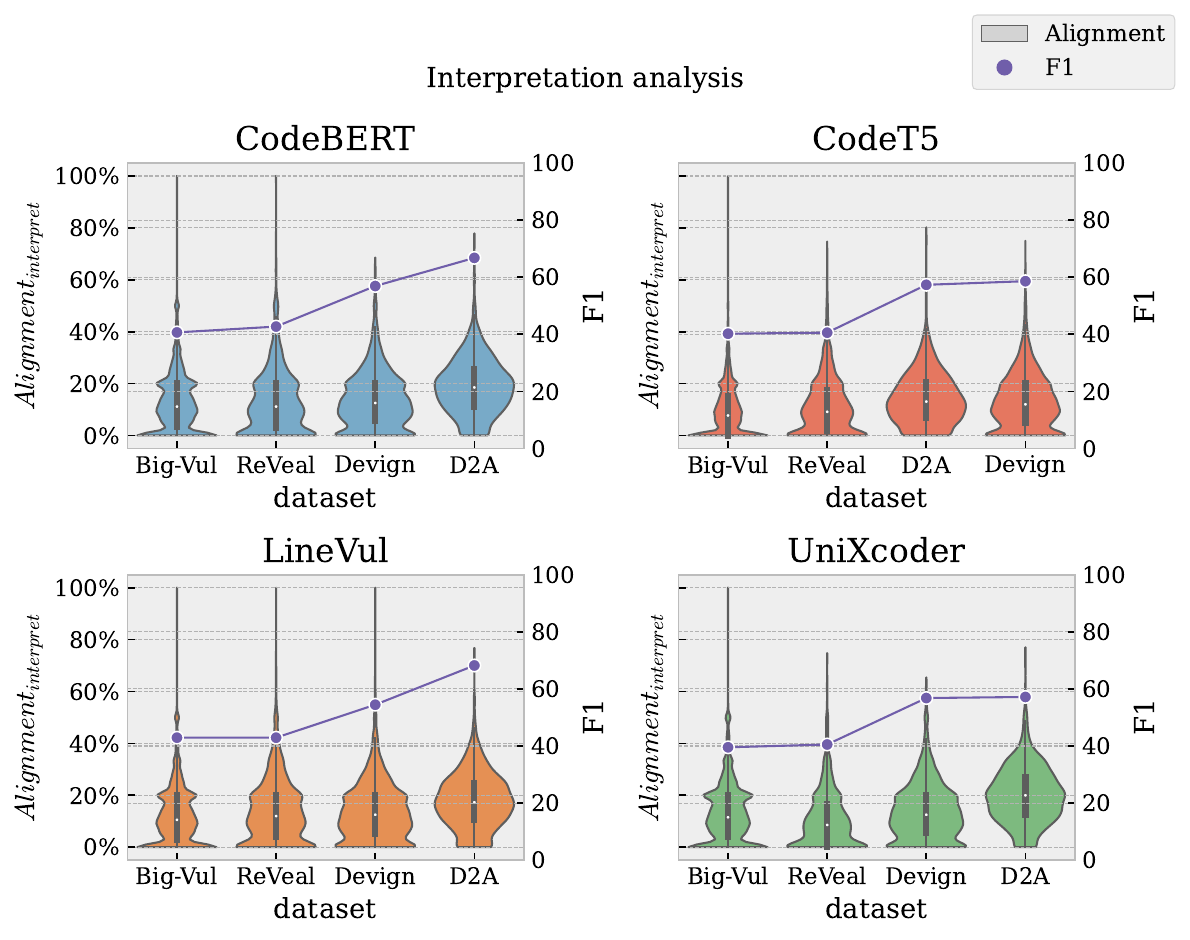}
\caption{Alignment of model interpretations to PVS bug semantics; datasets are ordered by model F1 score.
}
\label{fig:interpretation-analysis-results}
\end{figure}

Figure \ref{fig:interpretation-analysis-results} reports the $Alignment_{Interpret}$ scores of individual examples in each dataset, compared with PVS bug features.
The X-axis is sorted by model F1 score on the test dataset.

Within each model, in most cases, a higher performance on a dataset corresponds to a higher median $Alignment_{Interpret}$ score.
This suggests that higher-performing models learned to focus somewhat on individual bug features.
However, we note that all models' median alignments lie within 10-20\%, and that some datasets (e.g. Big-Vul) had many examples with 0\% alignment.
This indicates that the models often failed to focus on bug semantic features and rather focused on spurious features; we attempt to repair this misalignment by annotating the bug features in Section \ref{sec:improvement}.

Figure \ref{fig:d2a-bp} (left) shows the alignment of the model to the buggy paths in the D2A dataset.
All models had a median alignment score of 5\% or below, and there was no clear difference between different models' alignment; therefore, we conclude that the models weren't able to align to buggy paths.

\subsection{Attention analysis}

\subsubsection*{Motivation}
We analyze the transformer architecture's self-attention mechanism to provide an explanation of the model's prediction using methods different from interpretation tools. In contrast to gradient-based interpretation methods which are applied post-hoc, attention is the primary mechanism of transformer models, so it can be used as a direct explanation of the model's prediction.
In addition, the attention score signifies the connections between multiple locations. In future work, we plan to analyze the models' attention scores to understand whether the models can connect between multiple-location bug semantics.

\subsubsection*{Approach}

Intuitively, a high attention score from token $i$ to token $j$ causes the model to focus more on the previous layer's encoding of token $j$ as it encodes token $i$ \cite{alammar2018illustrated}.
Thus, if a layer has a high attention score between two tokens, then we conclude that the model has learned that these two tokens are important in relation to each other.
If one of the tokens is inside $B$, this is evidence that the model has learned part of the semantics of this bug.
To measure this evidence, we adapted the method established by Wan et al.~\cite{wanWhatTheyCapture2022} and Vig et al.~\cite{vigBERTologyMeetsBiology2021}.
The previous approach, namely Wan et al., directly compared the high-attention edges with edges in an AST.
Bug semantics are defined as sets of nodes, so we extend the approach to consider any node incident to any high-attention edge; this requires the model attention to align closely to the bug features.



We computed the IoU formula in Equation \ref{eq:attention-analysis} for each program, separately for each self-attention layer/head, where $\mi_{topAttention(k)}$ represents the set of $k$ tokens incident to the topmost attention scores, and $k=|B|$.

\begin{equation}\label{eq:attention-analysis}
Alignment_{Attention} =
    \frac{|\mi_{topAttention(k)} \cap \gt|}
    {|\mi_{topAttention(k)} \cup \gt|}
\end{equation}

In order to compare $\mi$ and $\gt$ at the same granularity, we averaged the attention scores incident to the tokens within each AST token.
Wan et al. applied a threshold to exclude any attention heads with fewer than 100 high-attention edges; because we compute the IoU, we disable this threshold.
\subsubsection*{Results}

\begin{figure}[t]
\centering
\includegraphics[width=0.49\textwidth]{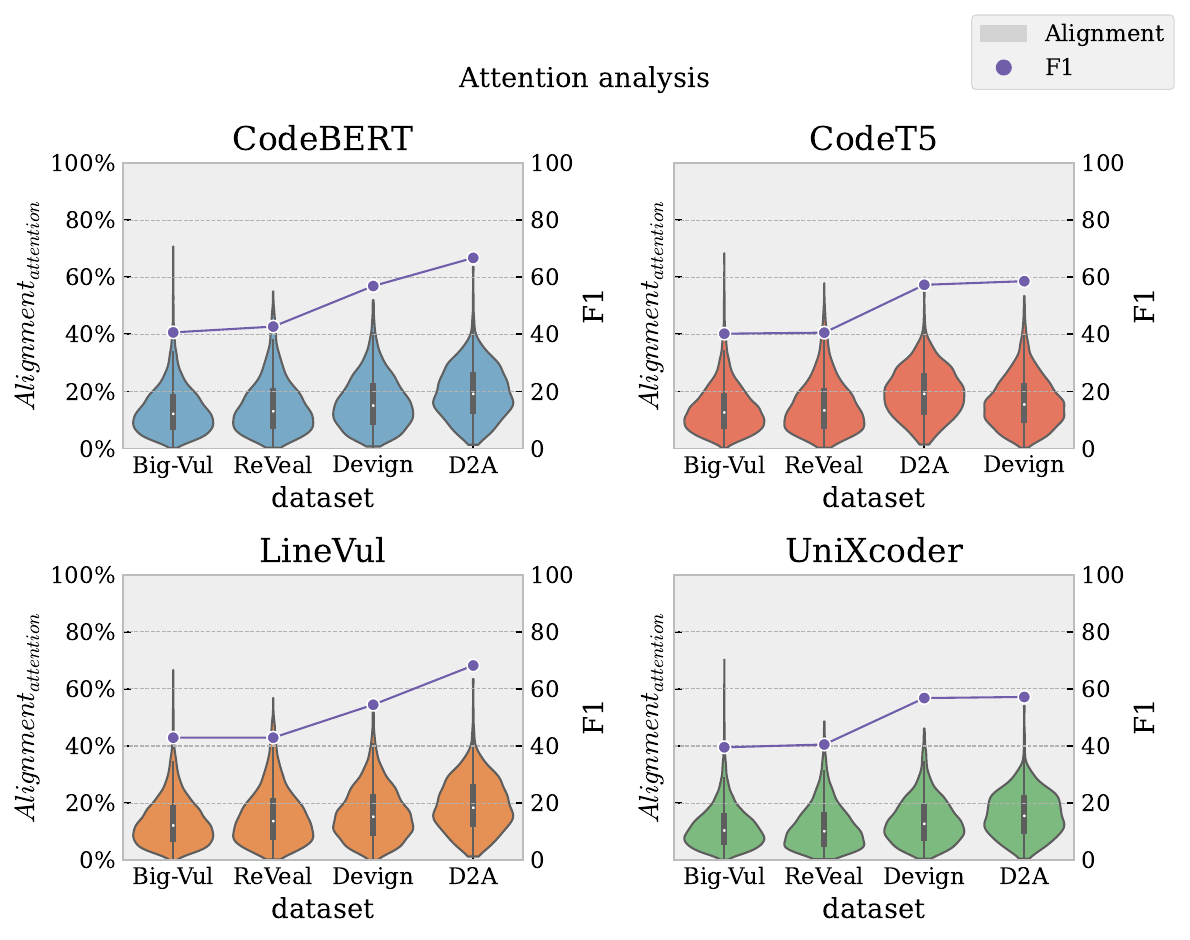}
\caption{Alignment of model attention to PVS bug semantics, aggregated by dataset example.}
\label{fig:attention-analysis-results}
\end{figure}

Figure \ref{fig:attention-analysis-results} reports the mean $Alignment_{attention}$ scores aggregated over the dataset.
%
%
The results of attention analysis corroborate the results of interpretation analysis.
We see that models tend to align better on better-performing datasets.
The median alignment scores for all models and datasets were between 10-20\%.
This shows that beyond highlighting the important features individually, the models also highlighted relationships between the important features using the attention mechanism.

\begin{figure}[t]
\centering
\includegraphics[width=0.49\textwidth]{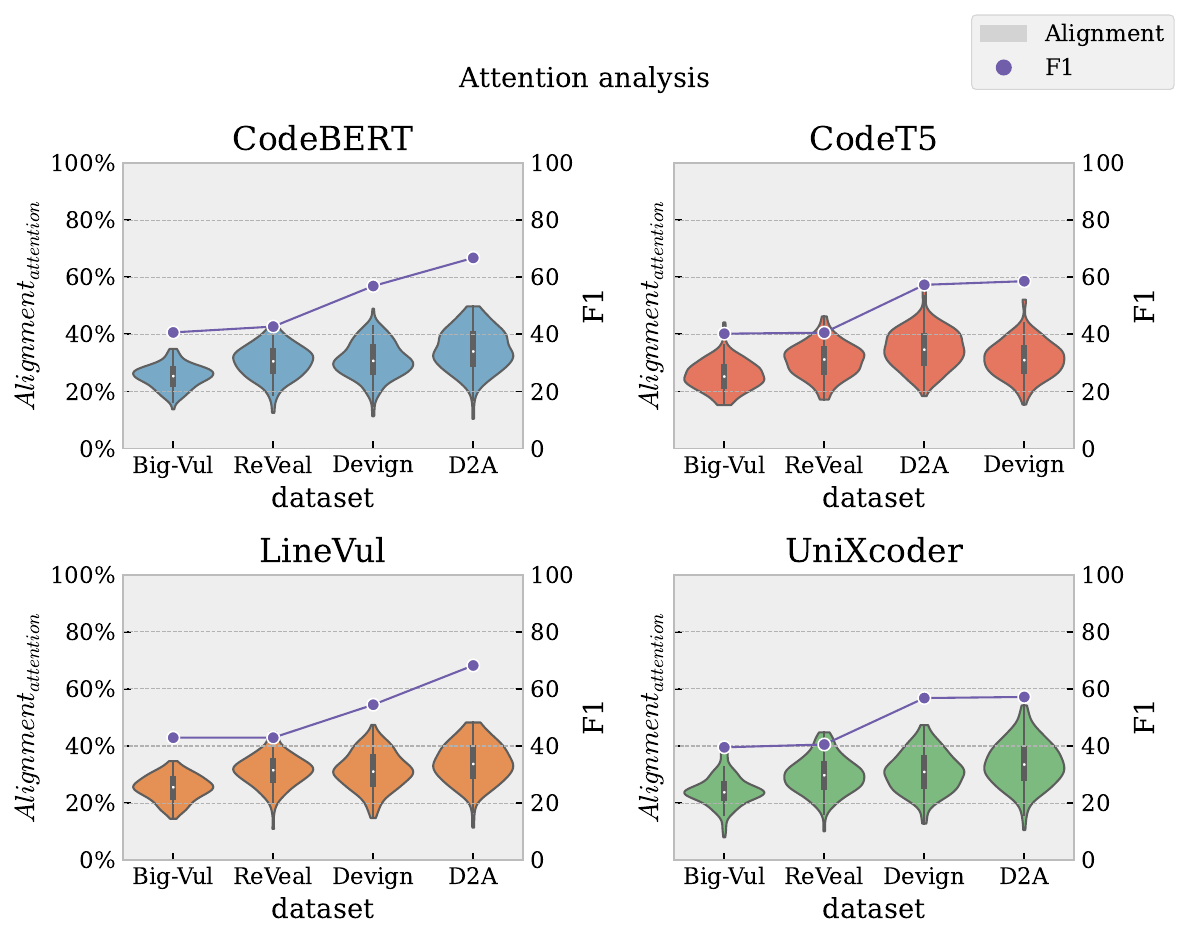}
\caption{Alignment of model attention to PVS bug semantics, aggregated by head.}
\label{fig:attention-analysis-results-by-head}
\end{figure}

Figure \ref{fig:attention-analysis-results-by-head} shows the results of the same attention scores, aggregated by head rather than by dataset example.
As the attention analysis computes scores per self-attention head and layer, we can observe different ``specialized'' parts of the same model by considering one attention head at a time.
All models, on all datasets, had some heads with low alignment (15\%), indicating that not all attention heads aligned to bug semantics; these heads may attend to other relationships, such as syntactic relationships in the AST~\cite{wanWhatTheyCapture2022}, delimiters, or the relative positions of tokens~\cite{clarkWhatDoesBERT2019b}.
A few attention heads had relatively high alignment scores -- up to 40\% in the case of the Big-Vul dataset and 50\% in the case of the D2A dataset.
On the other hand, most attention heads were clustered in a bell curve around 20-30\% alignment, indicating that most parts of the model did not focus solely on bug semantics.

Similar to interpretation analysis, the models all had low alignment to buggy paths (Figure \ref{fig:d2a-bp}, middle), showing that while models attended to potentially vulnerable statements, they did not learn the significance of buggy paths.

\subsection{Interaction Matrix}

\subsubsection*{Motivation}


The previous approaches analyzed the effects of individual tokens and connections between tokens, but did not consider the order of computation in the model.
Transformer models compute high-level representations of their sequences and their focus can move from one token to another throughout their computation.
Paltenghi et al.~\cite{paltenghiExtractingMeaningfulAttention2022} proposed the interaction matrix to represent this navigation of a model through the source code.
In order to provide a third distinct perspective on explaining the model, we used the interaction matrix to analyze the model.

\subsubsection*{Approach}

We computed an \textit{interaction matrix}, defined as $\im \in V \times V$. It measures which token will likely be attended by the model after the current token.

Concretely, each cell of $\im$ assigns a probability $\im(v_i, v_j)$ to each pair in $V \times V$ which represents the probability that the node in position $j$ will be focused on the model soon after the node in position $i$.
The authors originally used the interaction matrix to compare the attention of large language models with the code navigation of human developers.
We repurposed this analysis method in order to understand whether the model is likely to jump to or from bug semantic features when predicting an example.
We extended the definition to bidirectional models since we studied models with the BERT architecture, which attends to both future and past tokens, unlike GPT-like (decoder-only) models, and rather than applying to the generation task, we applied to the binary classification task.

To corroborate our results with the previous two metrics, we generated the interaction matrix $IM$, then computed the IoU metric on the edges with the highest probability as estimated by $IM$.
We list the formula in Equation \ref{eq:interaction-matrix}, where $\mi_{topIM(k)}$ is the set of $k$ tokens incident to highest transition probabilities estimated by the interaction matrix, and $k = |B|$.

\begin{equation}\label{eq:interaction-matrix}
Alignment_{IM} =
    \frac{|\mi_{topIM(k)} \cap \gt|}
    {|\mi_{topIM(k)} \cup \gt|}
\end{equation}

We averaged the probabilities in the interaction matrix within each AST token in order to compare $M$ and $B$ at the same granularity.

\subsubsection*{Results}

\begin{figure}
\centering
\includegraphics[width=0.49\textwidth]{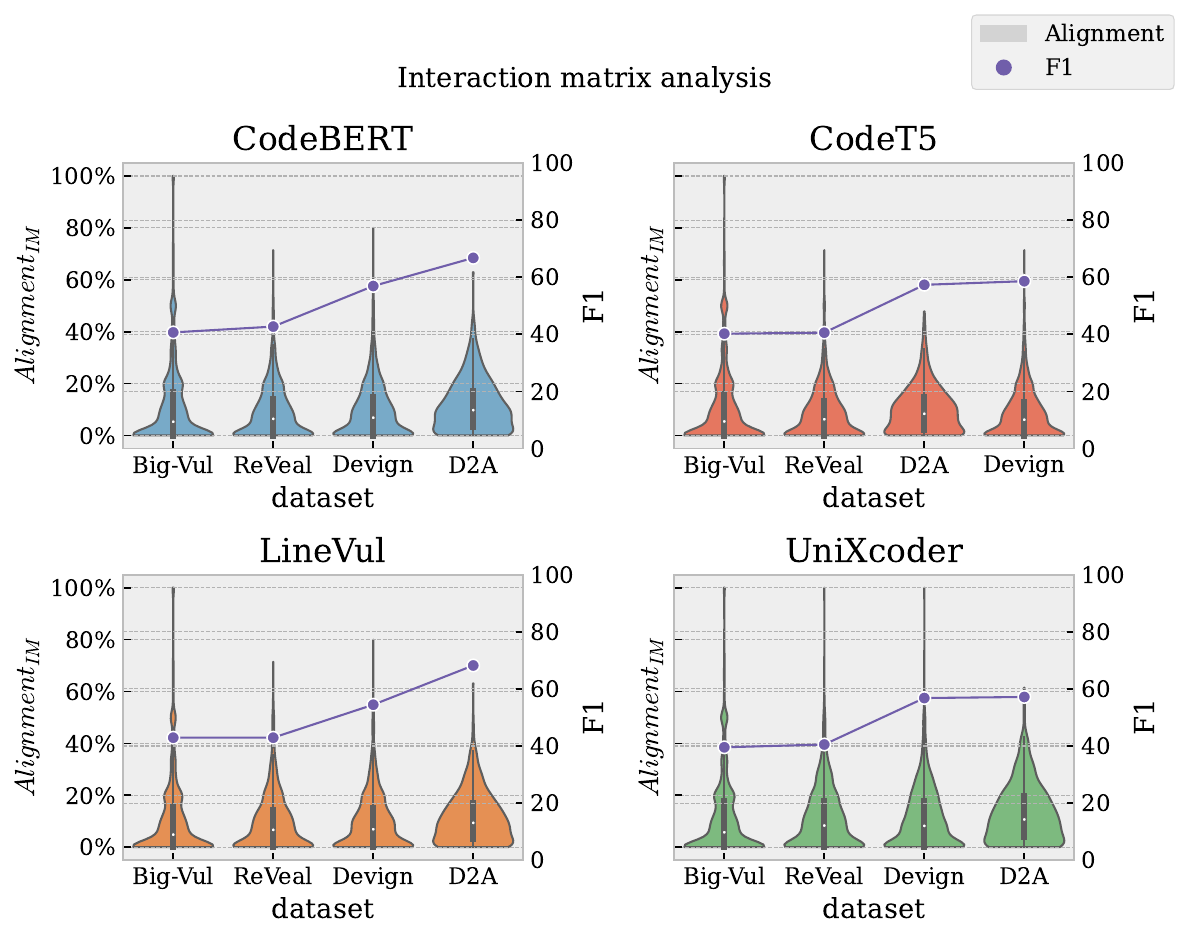}
\caption{Alignment of interaction matrix to PVS bug semantics.}
\label{fig:interaction-matrix-results}
\end{figure}

Figure \ref{fig:interaction-matrix-results} reports the $Alignment_{IM}$ scores from each example in the dataset, compared with PVS bug features.

The results of the interaction matrix corroborate the results of the interpretability and attention analysis, where tools which earned higher performance also aligned better on more examples.
The overall scores for the interaction matrix were lower than the previous approaches, as the highest median score of $Alignment_{IM}$ was around 10\% while the highest median score of $Alignment_{Interpret}$ was around 20\%.
This indicates that while the models could use the individual bug features for their predictions and relate the bug features within some attention heads, they were not likely to focus on the bug features throughout the computation of the model. The models may be focusing on some specific features which are common to PVS, without relating them to the context specific to the program.

The models all had relatively low alignment to buggy paths using the interaction matrix (Figure \ref{fig:d2a-bp} (right)).

\subsection{Findings from Alignment Studies: a Summary}





\paragraph{Finding 1: Better-performing models aligned better to PVS}
All models showed the lowest alignment with the Big-Vul dataset; the models demonstrated higher alignment with the balanced Devign and D2A datasets compared to the imbalanced Big-Vul and ReVeal datasets.
This trend matches the performance as well, with models performing worse on the imbalanced datasets than balanced datasets.
Imbalanced datasets have a lower ratio of vulnerable to non-vulnerable examples; this makes it more challenging for the model to learn to distinguish buggy from non-buggy code and may cause the models to focus less on (bug semantic) features which occur in buggy code~\cite{chakraborty_reveal_2022}.
However, we expect to see an imbalanced label distribution in real-world code~\cite{fan_msr_bigvul_2020}, so we hope to improve the models' ability to focus on bug features, even in an imbalanced setting.

\paragraph{Finding 2: The models did not align strongly to PVS}

Though the models did focus somewhat on PVS, the absolute scores of alignment were almost all below 50\%, indicating that the most influential features and attention scores did not align strongly with PVS.
We posit that if the models would align to the PVS more strongly, they would exhibit improved performance.
To this end, in the next section, we upgraded the models to make them focus more strongly on the PVS.


\paragraph{Finding 3: The models did not align to buggy paths in the D2A dataset}

Shown in Figure \ref{fig:d2a-bp}, all models failed to align to the buggy paths reported in D2A.
These buggy paths indicate the bug semantic features more precisely than PVS because they are constrained by a sequence of statements instead of one statement.
The lack of alignment indicates that the models did not learn these more precise bug semantics and opens up this area for future work.



\begin{figure*}[t]
    \newcommand{\tok}[1]{\texttt{#1}\,}
    \newcommand{\bmarker}{\textcolor{red}{\textsuperscript{begin}}}
    \newcommand{\emarker}{\textcolor{blue}{\textsuperscript{end}}}
    \centering
    %
    \renewcommand{\arraystretch}{2.0}
    \begin{tabular}{rl}
        \textbf{Original program}: \tok{[BOS]} & \tok{int} \tok{main} \tok{()} \tok{\{} $\underbrace{\tok{mall} \tok{oc} \tok{(} \tok{10} \tok{);}}_{\text{\makebox[0pt]{Potentially Vulnerable Statement (PVS)}}}$ \tok{\}} \tok{[EOS]} \\ 
        \textbf{Mark}: \tok{[BOS]} & \tok{int} \tok{main} \tok{()} \tok{\{} $\underbrace{\bmarker{}\tok{mall}\emarker{} \bmarker{}\tok{oc}\emarker{} \bmarker{}\tok{(}\emarker{} \bmarker{}\tok{10}\emarker{} \bmarker{}\tok{);}\emarker{}}_{\text{Inserted markers}}$ \tok{\}} \tok{[EOS]} \\ 
        \textbf{Prepend}: \tok{[BOS]} $\underbrace{\textcolor{red}{\tok{mall} \tok{oc} \tok{(} \tok{10} \tok{);}}}_{\text{Prepended PVS}}$ \textcolor{red}{\tok{[SEP]}} & \tok{int} \tok{main} \tok{()} \tok{\{} $\tok{mall} \tok{oc} \tok{(} \tok{10} \tok{);}$ \tok{\}} \tok{[EOS]}
    \end{tabular}
    %
    \caption{Examples of PVS annotations.}
    \label{fig:pvs-examples}
\end{figure*}

\begin{figure*}[b]
\centering
\includegraphics[width=0.33\textwidth]{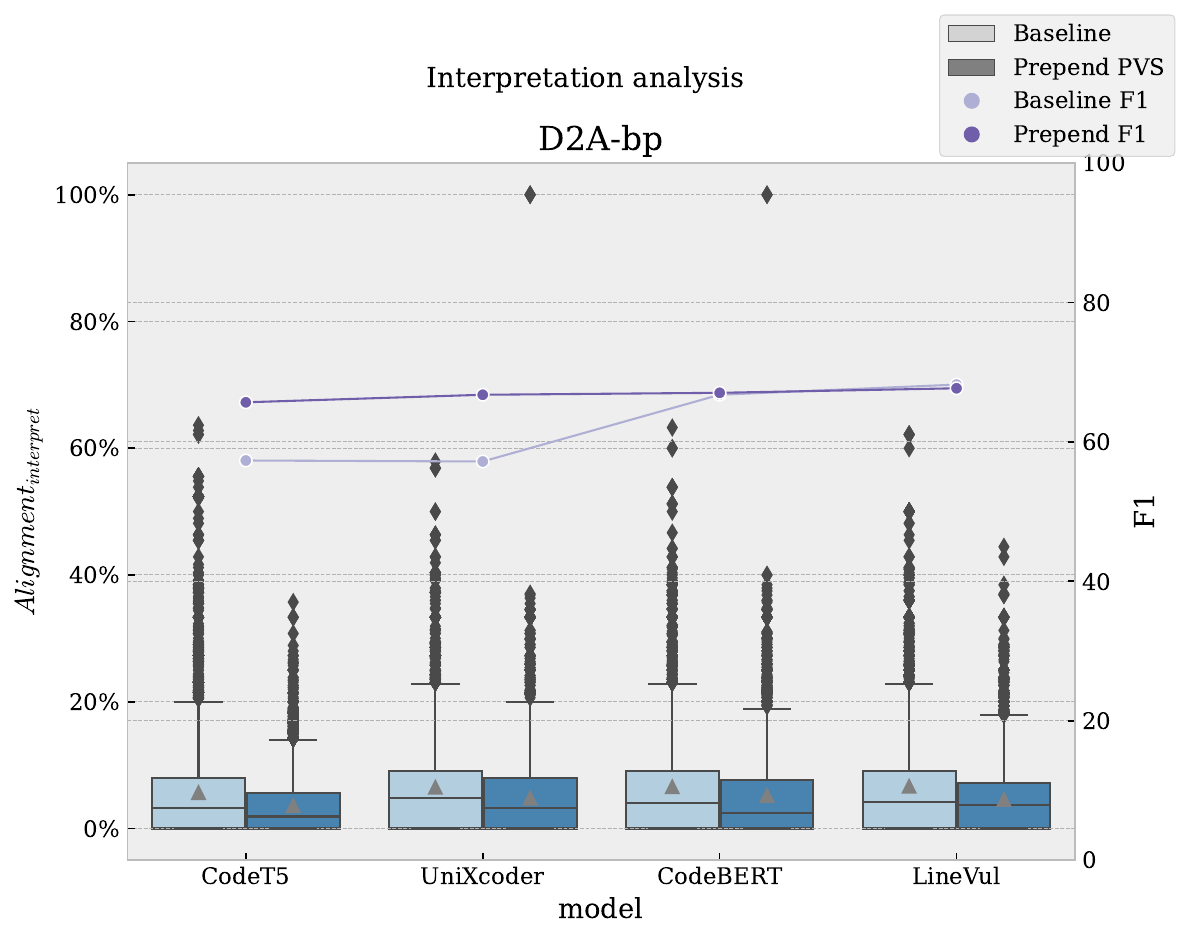}
\includegraphics[width=0.33\textwidth]{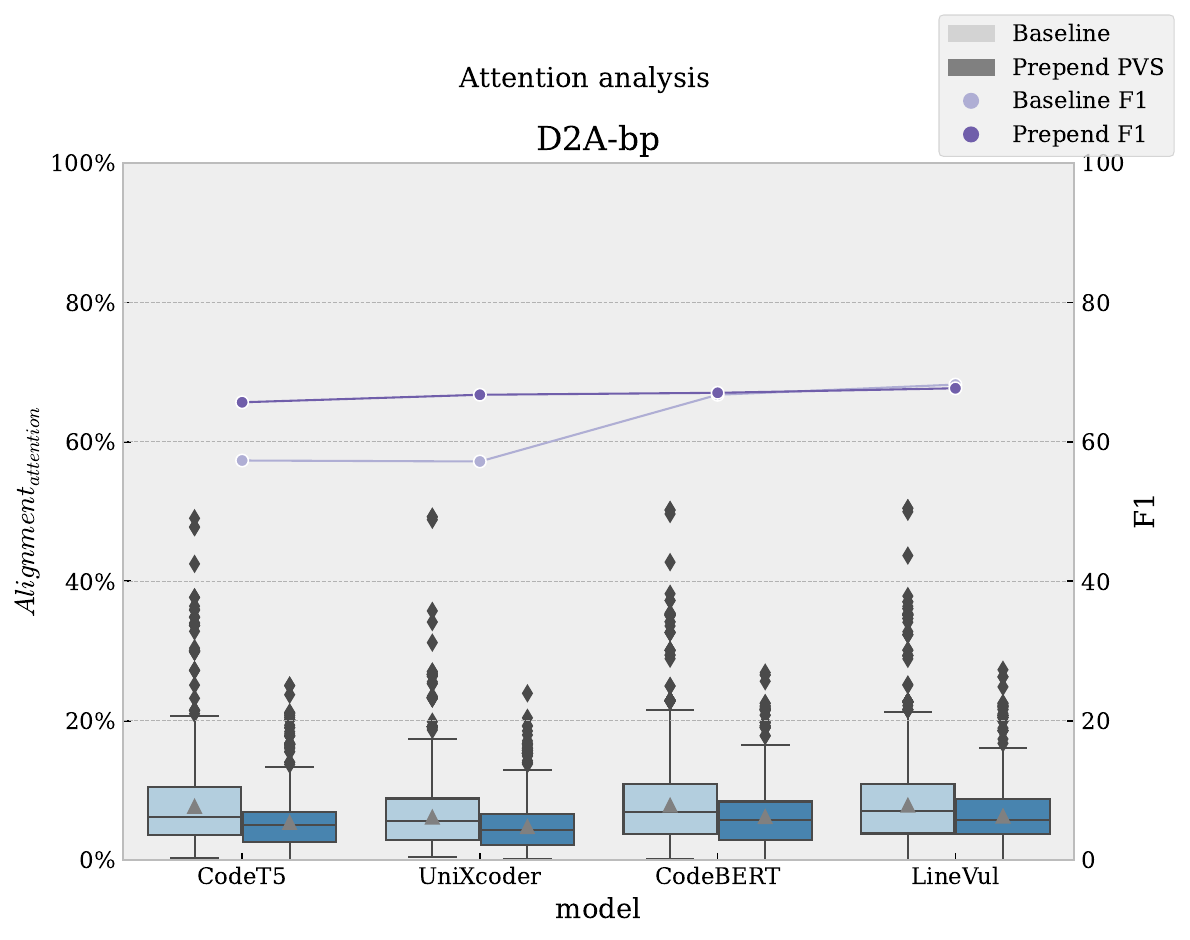}
\includegraphics[width=0.33\textwidth]{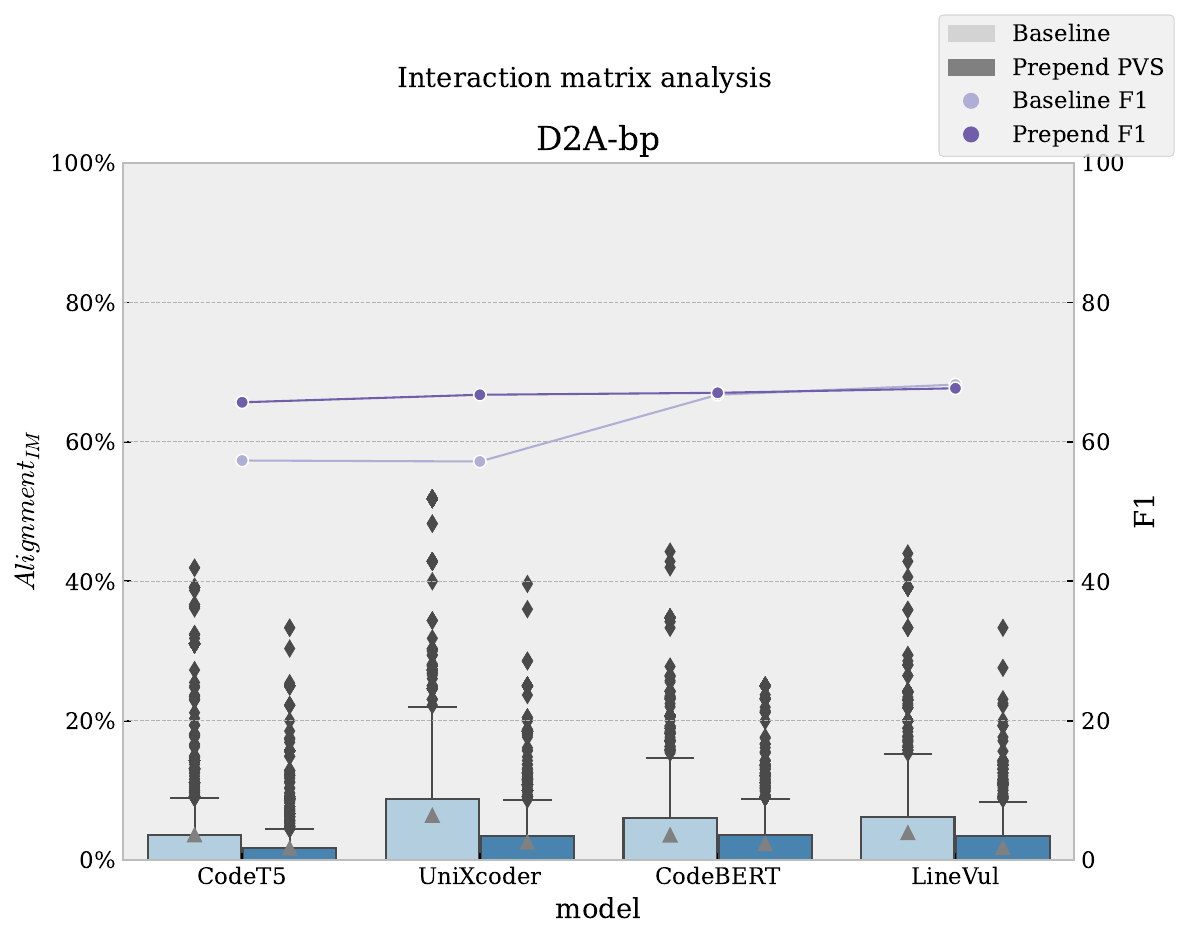}
\caption{PVS-annotated models did not align to D2A buggy paths.
}
\label{fig:pvs-d2a-bp}
\end{figure*}

\section{Improve the Models Using Bug Semantics}
\label{sec:improvement}

Based on the results of the alignment with PVS, we saw that the models' most important features aligned slightly with PVS, and that higher alignment was related to higher performance across different datasets.
However, the models' alignment scores were relatively low -- mostly below 50\% alignment.
We hypothesized that improving alignment to PVS would improve the performance of the model. To achieve the improved alignment, we annotated the PVS in the model inputs, and then study how the model performance changes.

\subsection{Annotating PVS in model inputs}

Several prior approaches have annotated language model input to give guidance or extra information to the model~\cite{SequenceR,chakraborty2021multimodal,chakrabortyCODITCodeEditing2022}.
We propose to annotate the Potentially Vulnerable Statements (PVS) inside the program in order to make the model more strongly focus on these for its predictions. This can be considered as inserting domain knowledge into the model's input.
%
Due to the large design space of the annotations, we developed several approaches and narrowed down to the two highest-performing annotation methods. 
These approaches are pictured in Figure \ref{fig:pvs-examples}, in contrast to the original example program in the top row.

\paragraph{Mark}
Inspired by the usage of boundary tokens in the original BERT model, we insert special ``marker'' tokens (\textcolor{red}{begin}, \textcolor{blue}{end}) before and after each BPE token inside the PVS~\cite{devlinBERTPretrainingDeep2019}. 
The marker tokens allow the model to use its positional encoding to attend to the tokens which are close to the markers~\cite{clarkWhatDoesBERT2019b}.
Figure \ref{fig:pvs-examples} (middle) shows the \textcolor{red}{begin} and \textcolor{blue}{end} tokens inserted around each token in the PVS. To represent these new tokens, we inserted new randomly initialized indices into the model's embedding matrix and trained the new indices during fine-tuning.
Since the model input tokens are broken up by the BPE tokenizer, this method tends to insert multiple marker tokens per statement.

\paragraph{Prepend}
Inspired by the hints provided to the MODIT model proposed by Chakraborthy et al.~\cite{chakraborty2021multimodal}, we prepend the tokens inside PVS at the beginning of the code.
Figure \ref{fig:pvs-examples} (bottom) shows that the original program's source code is unchanged, but we prepend the sequence \textcolor{red}{\texttt{malloc(10); [SEP]}} before the program's source code. \texttt{[SEP]} is simply a separator token used to separate the two semantically different sequences.
The placement of the PVS as a different modality before the input sequence allow the model to easily focus on these important statements~\cite{chakraborty2021multimodal}
Due to the models' limited context length of 512 tokens~\cite{feng_codebert_2020}, we truncated the prepended sequence to the first 100 BPE tokens.

\subsection{Results}

\subsubsection*{Performance}

\begin{table}[]
\centering
\caption{PVS annotations improved model performance in 56\% of models and datasets.}
\label{tab:pvs-performance}
\begin{tabular}{ll cccc}
\hline
\thead{Model}        & \thead{Setting} & \thead{D2A} & \thead{Devign}       & \thead{Big-Vul} & \thead{Reveal} \\ \midrule
CodeBERT  & Baseline & 66.76 & 56.90 & \textbf{40.65} & 42.69 \\
CodeBERT  & Mark  & 58.69 & \textbf{59.04} & 37.96 & 37.27 \\
CodeBERT  & Prepend  & \textbf{67.04} & 53.46 & 38.14 & \textbf{43.66} \\\midrule
UniXcoder & Baseline & 57.19 & 56.81 & 39.55 & 40.53 \\
UniXcoder  & Mark  & 60.10 & 59.56 & \textbf{39.63} & \textbf{43.67} \\
UniXcoder & Prepend  & \textbf{66.76} & \textbf{62.04} & 39.12 & 42.67 \\ \midrule

CodeT5    & Baseline & 57.33 & 58.79 & 40.20 & 40.56 \\
CodeT5  & Mark  & 52.63 & 4.67 & 39.78 & 42.73 \\
CodeT5    & Prepend  & \textbf{65.68}                   & \textbf{60.53} & \textbf{40.96} & \textbf{43.88} \\\midrule
LineVul   & Baseline & \textbf{68.22} & 54.15 & \textbf{39.46} & 42.92 \\
LineVul  & Mark  & 63.49 & 58.78 & 32.69 & 42.89 \\
LineVul   & Prepend  & 67.69 & \textbf{58.66} & 36.44 & \textbf{45.45} \\ \bottomrule
\end{tabular}
\end{table}

In Table \ref{tab:pvs-performance}, we present the performance of PVS annotations in three different settings.
\textit{Baseline} denotes the original model input, without changes. \textit{Mark} and \textit{Prepend} denote our proposed approaches.

The Prepend annotation increased model performance above the baseline in 11/16 cases; Mark did not perform as well, improving only 7/16 cases.
In fact, on the Devign, D2A, and ReVeal datasets, the Prepend annotation improved the performance of three out of four models.
This shows that, while it does not improve all cases, annotating the bug semantics can help model performance in the majority of cases.
In the best case (UniXcoder/Prepend on D2A), the annotation improved the F1 score by 9.57 points above the baseline.

However, in some cases such as CodeT5/Mark on Devign, the model performance reduces or collapses.
This may be caused by excessive annotations replacing some of the original code tokens in the model input, thus causing the signal to be unstable during training; since the insertion of annotations requires truncation of the original code tokens, this is a tradeoff due to the limitations of current model architectures.
However, newer models support a larger context length, which may allow more flexible and stable performance from annotation schemes.
Based on Table \ref{tab:pvs-statistics}, the MSR dataset has the fewest PVS per program; this may contribute to reducing the performance of the annotation method. Future work should seek to improve this limitation by providing a more robust annotation method or including more function calls in the list of PVS.


\subsubsection*{Alignment}

\begin{figure}[t]
\centering
\includegraphics[width=0.49\textwidth]{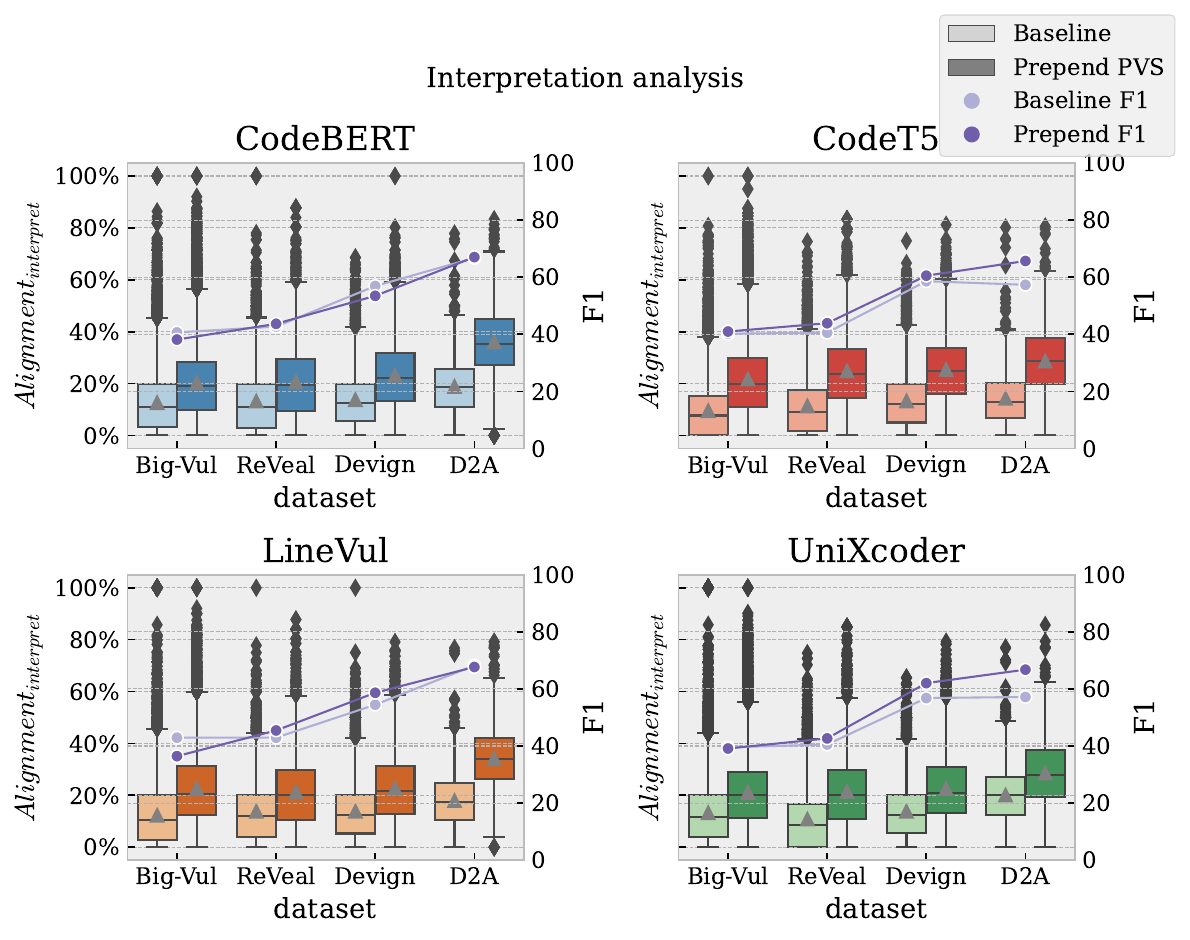}
\caption{Alignment of baseline vs. annotated model interpretations to PVS bug semantics.}
\label{fig:interpretability-analysis-results-pvs}
\end{figure}

\begin{figure}[t]
\centering
\includegraphics[width=0.49\textwidth]{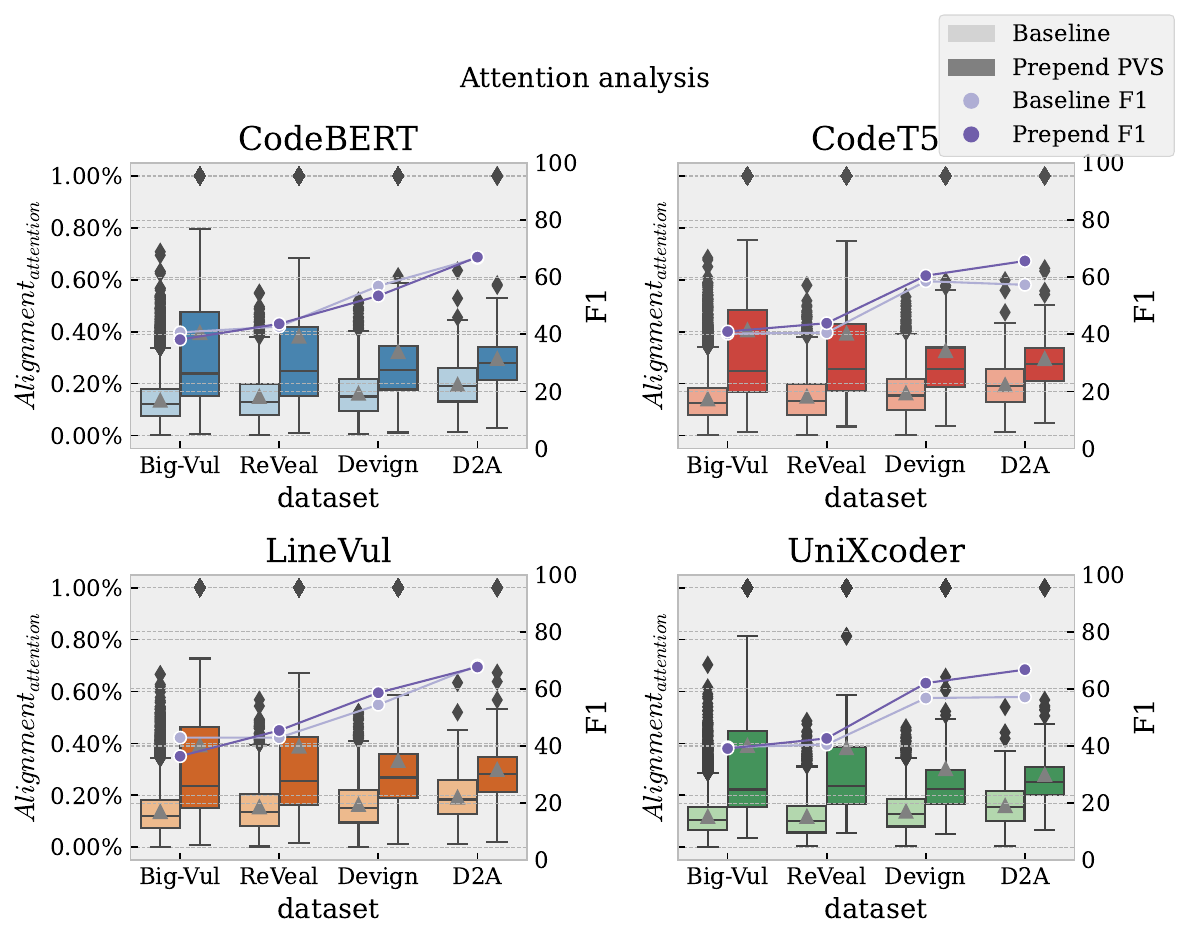}
\caption{Alignment of baseline vs. annotated model attention to PVS bug semantics, aggregated over all attention heads by dataset example.}
\label{fig:attention-analysis-results-pvs}
\end{figure}

\begin{figure}[b]
\centering
\includegraphics[width=0.49\textwidth]{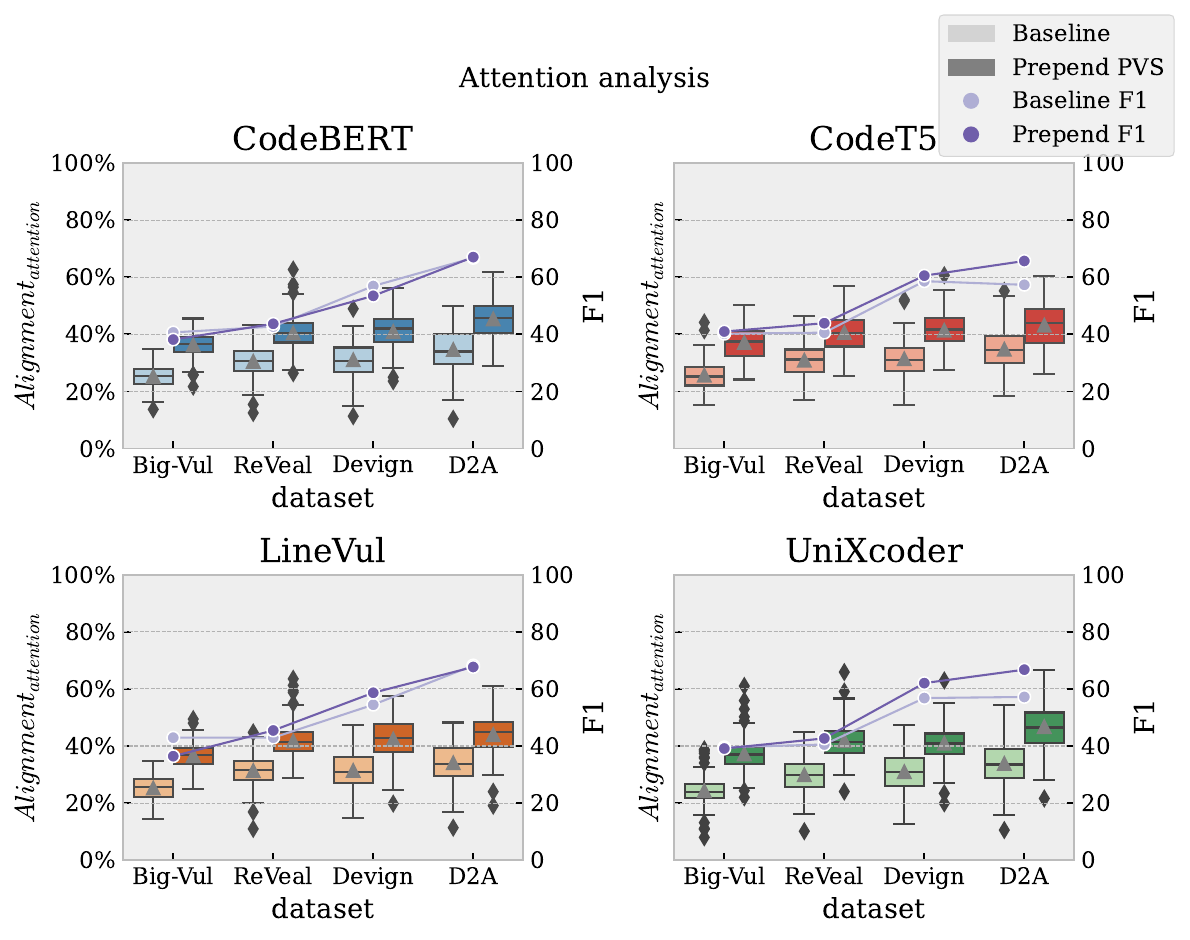}
\caption{Alignment of baseline vs. annotated model attention to PVS bug semantics, aggregated by attention head.}
\label{fig:attention-analysis-results-pvs-by-head}
\end{figure}

\begin{figure}[t]
\centering
\includegraphics[width=0.49\textwidth]{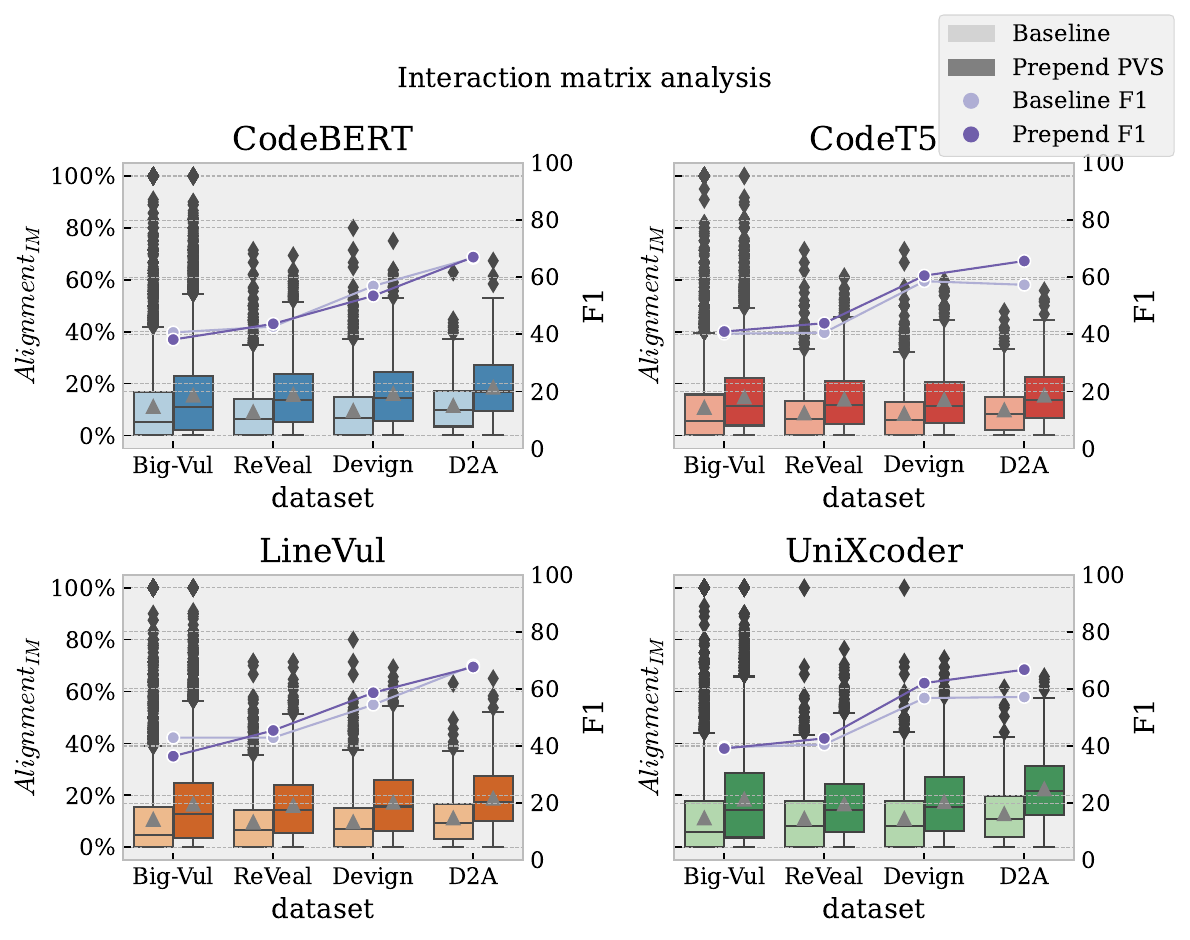}
\caption{Alignment of baseline vs. annotated model interaction matrix to PVS bug semantics.}
\label{fig:interaction-analysis-results-pvs}
\end{figure}

We measured the alignment on the best-performing approach, Prepend.
Since the Prepend approach copy-pastes the PVS from the code, we considered all the tokens in the prepended annotations as part of the buggy set $B$.
Figures \ref{fig:interpretability-analysis-results-pvs}, \ref{fig:attention-analysis-results-pvs}, and \ref{fig:interaction-analysis-results-pvs} show the results; the Baseline and Prepend performance are overlaid using dotted lines.
The PVS annotation increased the models' alignment scores on all models and datasets.
The mean alignment increased from 36-111\% on interpretability analysis, 51-232\% on attention analysis, and 38-79\% on interaction matrix analysis.
This is strong evidence that the models learned to use bug semantic features from the annotated inputs to make their predictions.
Investigating further, Figure \ref{fig:attention-analysis-results-pvs-by-head} shows the alignment scores aggregated by each attention head across all examples.
\textit{All} parts of the model learned to align with PVS, evidenced by the higher minimum and maximum alignment scores.
Additionally, the model still learned to ``specialize'' with some attention heads learning to focus on PVS, up to 65\% alignment score.
However, in some cases, such as CodeBERT, the PVS annotation failed to increase the model performance, even though alignment to PVS increased. This confirms our original idea that alignment to PVS is not sufficient to detect all bugs.

Next, we measured alignment of the annotated model with D2A buggy paths. Figure \ref{fig:pvs-d2a-bp} shows that the PVS annotations reduced the baseline model's alignment to buggy paths.
It is unclear why this occurred; but we plan to further study the relationship between PVS and buggy paths in future work.

We note that the annotations we proposed are only a prototype, and further development of this idea could further improve the alignment and/or performance. We leave such improvements to future work.
We also tried to annotate the PVS using Highlight-Transformer~\cite{liuHighlightTransformerLeveragingKey2021}, which assigns higher attention weights to special tokens, but couldn't increase the performance with this approach.
In addition to annotating the PVS, we have also tried to annotate path-based bug semantics by annotating statically detectable source/sink bug features, but in our initial experiments, the simple annotation technique couldn't increase the performance in most cases.
We view alignment to path-based bug semantics as an important next step and encourage future work in this area.

\section{Threats to Validity}

There have been many techniques which attempt to understand model behaviors, but evaluation of these techniques often faces challenges and has limitations. To increase the reliability of our conclusions, we corroborated three distinct techniques -- interpretability tools, attention analysis, and interaction matrix analysis -- each of which provides different perspectives of the model.

We used interpretability tools, which may not faithfully estimate the attributions of features. To mitigate this concern, we aggregated the attribution scores from multiple tools which used distinct techniques.
%
Furthermore, interpretations of the attention score do not fully explain model predictions -- they can only show correlation with high-attention edges~\cite{jainAttentionNotExplanation2019,wiegreffeAttentionNotNot2019}.
To understand the explanations of the model's predictions, we studied both interpretability tools and the attention mechanism.

We generated our list of potentially vulnerable statements from the C standard library combined with our domain knowledge. However, some function calls may be missing from the list of PVS, which may impact the measurement of alignment with bug semantics.




Finally, it would be interesting to study larger language models such as ChatGPT and GPT-4, which have exhibited superior capabilities to BERT family models on source code understanding tasks. However, these models are closed-source, so we are not able to use interpretability tools or analyze their attention scores in detail.

\section{Related Work}

\paragraph{Empirical studies of deep learning in SE}

Niu et al.~\cite{niu2023empirical} compared the performances of 19 pre-trained source code models on 13 different software engineering tasks.
Croft et al. \cite{croft2023data} analyzed the quality of the open-source datasets published in the SE literature and evaluated the performance after data cleaning.
Chakraborthy et al.~\cite{chakraborty_reveal_2022} studied the performance of deep learning models in a real-world vulnerability prediction scenario.
Steenhoek et al.~\cite{steenhoek2023empirical} explored the important and difficult features for deep learning models, and evaluated their performance under novel training data sizes, bug types, open-source projects.
In contrast to these studies, we went beyond performance metrics and compared the models' influential features with bug semantics.



\paragraph{Model guidance}

For the program repair task, MODIT~\cite{chakraborty2021multimodal}, CODIT~\cite{chakrabortyCODITCodeEditing2022}, and SequenceR~\cite{SequenceR} wrapped the buggy code with special tokens such as \texttt{<START\_BUG>}, \texttt{<END\_BUG>} to guide the model toward the buggy lines. Inspired by these approaches, we designed an approach to guide the model to focus on PVS.
Liu et al.~\cite{liuHighlightTransformerLeveragingKey2021} introduced Highlight-Transformer, which assigns greater attention weights to key phrases in order to improve document summarization.
We tried to directly apply their approach to assign higher weights to the PVS, but our naive approach did not improve the performance.

\paragraph{Model analysis}

Various works have analyzed self-attention weights to understand the interaction between attention and the structure of the data.
Wan et al. \cite{wanWhatTheyCapture2022} conducted attention analysis to investigate how attention aligns with the syntactic structure of source code.
Paltenghi et al. \cite{paltenghiExtractingMeaningfulAttention2022} studied developer's and human's attention during code exploration. They proposed an interaction matrix from the attention weights of the models which captures the likelihood of the model's focus moving from one line to another.
We extended these methods to explain how the transformer models learn bug semantics.

\section{Conclusions and Future work}

We studied the alignment between influential model features, defined by interpretation tools, attention analysis, and interaction matrix analysis, and bug features, defined by Potentially Vulnerable Statements (PVS) and buggy paths.
We show that the better-performing models aligned better to PVS; however, the models did not strongly align to PVS. Additionally, the models did not align substantially to buggy paths.
Based on our analysis, we developed two methods for annotating the PVS inside the model inputs. We showed that the annotations improved the performance and alignment in the majority of settings; annotated models improved their F1 score by up to 9.57 points and aligned up to 232\% better to PVS.

Our annotation approach is a preliminary attempt at using PVS to improve the model performance and alignment. In the future, we will also develop other approaches to make the models attend to bug semantics. Another highly-motivated future work is to study the alignment to path-based bug semantics rather than comparing individual statements and to make the model align to buggy paths in order to improve its performance.
These can include static analysis paths as well as execution traces.




\bibliographystyle{ACM-Reference-Format}
\bibliography{main}



\end{document}